# Deep Learning in Early Alzheimer's Disease's Detection: A Comprehensive Survey of Classification, Segmentation and Feature Extraction Methods

[1]Rubab Hafeez, [1]Sadia Waheed, [1]Syeda Aleena Naqvi, [2]Fahad Maqbool, [1]Amna Sarwar, [3]Sajjad Saleem, [4]Muhammad Imran Sharif, [5]Kamran Siddique and [6]Zahid Akhtar

[1]*Department of Computer Science, University of Wah, Wah Cantt, Pakistan*
[2]*School of Electrical Engineering and Computer Sciences (SEECS), NUST, Pakistan*
[3]*Department of Information and Technology, Washington University of Science and Technology Virginia, Virginia, USA*
[4]*Department of Computer Science, Kansas State University, Manhattan, Kansas, USA*
[5]*Department of Computer Science and Engineering, University of Alaska Anchorage, Anchorage, USA*
[6]*Department of Network and Computer Security, State University of New York Polytechnic Institute, USA*

Corresponding Author:
Rubab Hafeez
Department of Computer Science, Air University Aerospace and Aviation Campus Kamra
Email: rubab.rhk@gmail.com

**Abstract:** Alzheimer's disease is a deadly neurological condition, impairing important memory and brain functions. Alzheimer's disease promotes brain shrinkage, ultimately leading to dementia. Dementia diagnosis typically takes 2.8-4.4 years after the first clinical indication. Advancements in computing and information technology have led to many techniques for studying Alzheimer's disease. Early identification and therapy are crucial for preventing Alzheimer's disease, as early-onset dementia hits people before the age of 65, while late-onset dementia occurs after this age. According to the 2015 World Alzheimer's Disease Report, there are 46.8 million individuals worldwide suffering from dementia, with an anticipated 74.7 million more by 2030 and 131.5 million by 2050. Deep Learning has outperformed conventional machine learning techniques by identifying intricate structures in high-dimensional data. Convolutional Neural Network (CNN) and Recurrent Neural Network (RNN) have achieved an accuracy of up to 96.0% for Alzheimer's disease classification and 84.2% for Mild Cognitive Impairment (MCI) conversion prediction. There have been few literature surveys available on applying ML to predict dementia, lacking in congenital observations. However, this survey has focused on a specific data channel for dementia detection. This study evaluated deep learning algorithms for early Alzheimer's disease detection using openly accessible datasets, feature segmentation, and classification methods. This article also has identified research gaps and limits in detecting Alzheimer's disease, which can inform future research.

## Introduction

Patients with Alzheimer's disease experience severe symptoms like memory and visual loss, speech difficulties, lack of motivation, difficulty making key decisions, and mood fluctuations over time (Agarap, 2018; Muhammed Raees and Thomas, 2021). Older individuals diagnosed with early Alzheimer's disease detection slows down its progression to spread (Blom *et al*., 2009). Few early stages of Alzheimer's disease have been detected using machine learning algorithms (Blom *et al*., 2009; Stopschinski *et al*., 2021).

Two of the most popular tests for assessing AD are the Mini-Mental State Examination (MMSE) (Stopschinski *et al*., 2021) and the Clinical Dementia Rating (CDR) (Belam and Nilforooshan, 2021); however, it should be highlighted that utilizing these tests as ground truth labels for AD may not be accurate. According to the previously stated criteria, clinical diagnosis of AD has been reported to have 70-90% accuracy rates when compared to post-mortem diagnosis (Littau, 2022 Ávi;la-Jiménez *et al.*, 2024). Clinical diagnosis is the best reference standard currently available despite its limitations (Sharma *et al.*, 2021). It's also important to remember that not all of the recognized biomarkers are readily available.

According to reports in 2010, 35.6 million people over 60 worldwide and 310,000 in Australasia were estimated to be suffering from dementia. It is anticipated that the population will nearly double every 20 years, with 790,000 people living in Australasia and 115 million people worldwide by 2050 (Ebrahimighahnavieh *et al.*, 2020). With 13,126 cases recorded in 2016, dementia has risen to the position of the second most common cause of mortality in Australia. One of the most expensive chronic disorders is Alzheimer's Disease (AD), and nursing care for people with AD and other dementias is predicted to rise significantly (Kalkan *et al.*, 2022).

Accurate dementia classification is challenging owing to variables including noisy MRI images and class imbalance issues in the multi-class categorization of AD, class imbalance could be resolved by adjusting the loss function to give more importance to the minority class which improves model sensitivity towards it, additionally, oversampling methods like Synthetic Minority Over-sampling Technique (SMOTE) create synthetic samples for the minority class, thereby balancing the class distribution. Combining predictions from multiple models or using techniques like boosting can improve performance on imbalanced datasets by aggregating the strengths of various models.

Alzheimer's disease affects almost 6.5 million Americans aged 65 and older (Ibrahim *et al.*, 2023). Between 2000 and 2019, more than 145% of diagnoses were made between 2.8 and 4.4 years following the onset of clinical symptoms (Ganesh *et al.*, 2023; Ahirwar, 2013). These symptoms indicate nerve cell damage in certain areas of the brain, which can lead to dementia. Dementia symptoms are categorized into three stages:

1) Initial stage
2) Mild stage
3) Severe stage

During the early phase, symptoms include agitation, suspicion, and the need for help with everyday duties. Dementia can affect a person's personality, causing them to be unaware of recent events and need round-the-clock care in the mild phase. In the severe or advanced stage, patients lose their memory, struggle with communication, recognize people, and require full assistance with daily activities (Pallawi and Singh, 2023; Keshri *et al.*, 2022; Arafa *et al.*, 2024).

Mild cognitive impairment is comparable to Alzheimer's disease (Pushpa *et al.*, 2019; Padmavathi *et al.*, 2023). The main contributions of the proposed survey are:

- It has provided a thorough analysis of historical machine-learning strategies for early detection of dementia
- It has comprehensively covered past classification, segmentation, and feature extraction approaches
- This study has highlighted problems and research needs in cutting-edge DL/ML approaches

The next section covers the architecture of the Deep Neural Network (DNN) model.

## *DNN Architecture and Transfer Learning Models*

To detect Alzheimer's disease, CNNs are particularly suited because:

- Feature extraction: CNNs are effective at automatically learning and extracting features from images, such as brain scans, which can include subtle patterns and anomalies indicative of Alzheimer's disease
- Hierarchical learning: The hierarchical nature of CNNs allows for capturing complex patterns at different levels of abstraction, which is crucial for identifying the progressive and nuanced features associated with Alzheimer's
- Previous success: CNNs have shown success in similar medical imaging tasks, such as detecting tumors or other neurological disorders, providing a precedent for their use in Alzheimer's detection

Deep neural networks have two main sets of layers in their design, the first one for detecting features and the other one for categorizing them. The network's feature detection layers perform data operations such as convolution, pooling, and ReLU. Rectified Linear Units (ReLU) is an activation function with strong biological and mathematical explanations that was first described by Hahnloser *et al.* in 2000. In 2011, it was demonstrated that it outperformed commonly used logistic sigmoid activation functions for training deep neural networks as of 2018 ReLU is one of the most popular activation functions for deep neural networks, (Agarap, 2018; Muhammed Raees and Thomas, 2021; Ibrahim *et al.*, 2023; Ganesh *et al.*, 2023).

ReLU works by thresholding values at 0. When $x_1<0$, it produces 0; otherwise, it produces a linear function on $x_1\geq 0$ shown in Eq. (1):

$$f\,(x_1) = x_+^1 = max(0, x_1) \quad (1)$$

Where $x_1$ is an input to the neuron, (Ahirwar, 2013), following the feature detection layers, the DNN architecture moves on to the classification layers. The FC layer generates a vector with $K$ dimensions, representing the total number of classes the network can predict. When it comes to this categorization difficulty, $K$ equals three. The softmax function generates a discrete probability distribution for $K$ classes (here, $K = 3$) shown in Eq. (2):

$$\sum_{k=1}^{K} P_k \quad (2)$$

Assuming $x_1$ as an activation function at the penultimate layer of the network and $\theta$ as a weight parameter, then input to the layer becomes Eq. (3):

$$S = \sum_{i}^{n-1} \theta_i x_i \quad (3)$$

And:

$$P_k = \frac{exp(S_k)}{\sum_{i}^{n-1} exp(S_k)} \quad (4)$$

And then the predicted class is shown in Eq. (5):

$$\hat{y} = argmax\, P_i; i \in 1,2,\ldots,N \quad (5)$$

Table (1) compares the outcomes of the proposed survey with subsequent analysis of Alzheimer's disease detection techniques which allows frameworks allows benchmarking against state-of-the art techniques, validation of the techniques and spotting the limitations.

### *Alzheimer's Disease and Non-Alzheimer's Disease Dementia*

Dementia is a memorial failing along with major life-altering mental impairment. There are several different kinds of dementia, with AD becoming the most prevalent. Since AD has evolved, there is now no cure. However, the precise cause is unknown; sickness is typically discovered during aging. To operate in various models, some fundamental procedures must be taken. We examine:

1) Preprocessing
2) Segmentation
3) Feature extraction
4) Classification after working through various articles

Figure (1) illustrates a holistic framework of the proposed survey.

We present the most recent findings and trends after carefully examining more than 100 publications in the literature. There are still numerous problems with deep learning, particularly in regard to the availability of training materials and data approaches, even though it has demonstrated a lot of success in detecting AD. A detailed comparison between existing techniques with respect to the limitations in predicting AD is depicted in Table (2).

**Table 1:** Comparison of the proposed framework with existing surveys

| Contents | Proposed survey | Pallawi and Singh (2023) | Keshri *et al.* (2022) | Blom *et al.* (2009) | Stopschinski *et al.* (2021) | Belam and Nilforooshan (2021) |
|---|---|---|---|---|---|---|
| Modalities | D | D | D | D | D | D |
| Performance measures | D | × | × | × | D | × |
| Open-source datasets | D | D | × | × | D | × |
| Research gaps | D | × | × | × | × | × |
| Limitations/ challenges | D | D | D | D | × | × |

**Table 2:** Limitations in existing findings

| Author | Adopted methodology | Findings (%) | Limitations |
|---|---|---|---|
| Littau (2022) | • SVM<br>• CNN | Accuracy: 75.4 | Methodology details are not included |
| Ebrahimighahnavieh *et al.* (2020) | Convolutional neural network | Accuracy: 93 | Complex medical images |
| Kalkan *et al.* (2022) | • CNN<br>• LDA | Accuracy: 84.2 | Mis-classification was observed in 3 class classifications |
| Pushpa *et al.* (2019) | • Decision tree<br>• SVM<br>• XGBoost<br>• Random forest | Accuracy: 86 | Failed to identify relevant attributes and features |
| Padmavathi *et al.* (2023) | • Random forest<br>• Classifier<br>• Decision tree<br>• Logistic regression<br>• Extra three classifier | Accuracy: 93 | Compromising algorithm's accuracy due to time-consuming calculations |

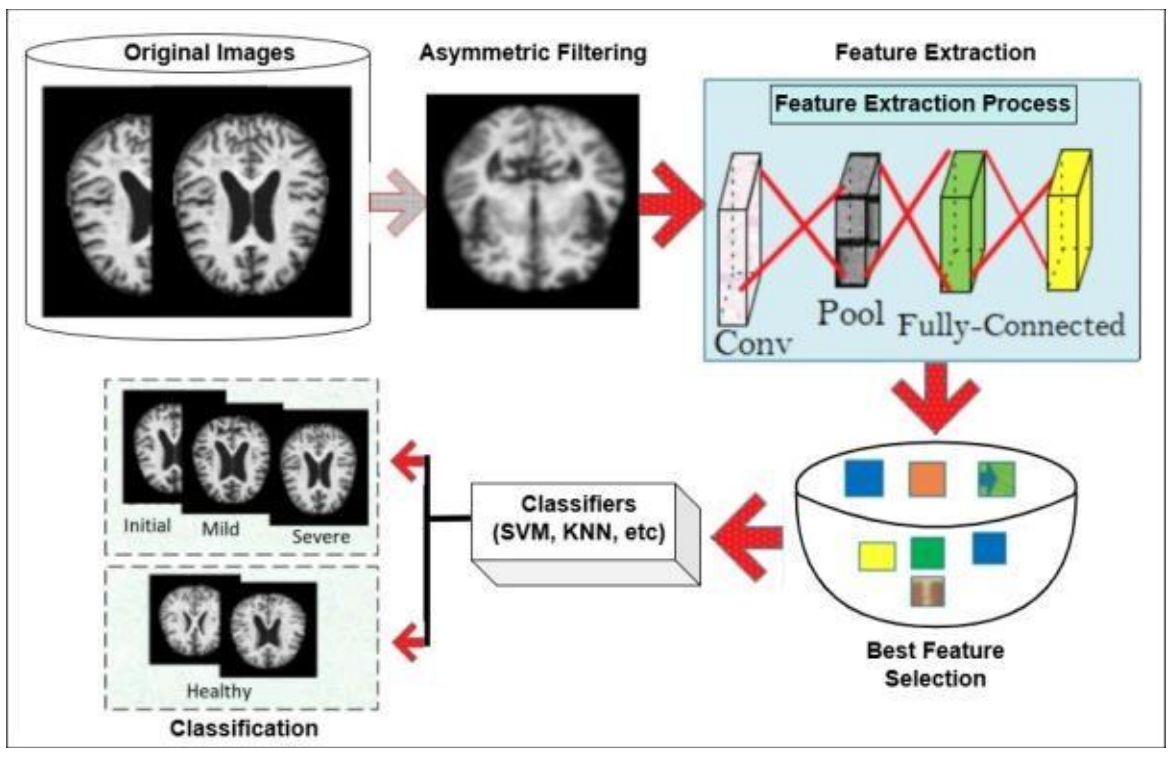

**Fig. 1:** Proposed framework

## Materials and Methods

### *Materials*

### *Available Datasets*

Alzheimer's disease patients are diagnosed using a three-dimensional (3D) cross-sectional brain MR imaging data processing approach that classifies several types of brain cells (GM, WM, and CSF). OASIS (Marcus *et al*., 2007), IBSR (Dayananda *et al*., 2022), Function Biomedical Informatics Research Network (FBIRN) (Keator *et al*., 2016), ADNI (Jack *et al*., 2008), MIRIAD and (Yiğit and Işik, 2020) e Kaggle's dataset contains 10432 pictures for testing connected to Alzheimer's disease. These photos are grouped into four categories.

- Class 0 "MildDemented"
- Class 1 "ModerateDemented"
- Class 2 "NonDemented"
- Class 3 "SevereDemented"

Table (3) shows the specifications of these datasets, there will be an evaluation of brain MRI at different phases.

### *Methods*

### *Preprocessing*

An image can be modified or have important information extracted from it using a technique called image processing. There are a number of artifacts in the image acquisition phase, including noise, fuzzy borders, blurriness, and skulls. Preprocessing techniques might be used to get rid of these artifacts. Preprocessing procedures are beneficial because of the bubbly borders, contrast, and noise in the MRI brain; this approach, which offers 89.99% accuracy on the OASIS dataset, transforms MRI pictures into grayscale images and applies histogram equalization for contrast improvement (Wulandari *et al*., 2018; Kumar *et al*., 2023). Participants in the OASIS dataset are frequently drawn from academic or research organizations, which may lead to selection bias in cases where the individuals are better educated and have greater access to healthcare than the overall population. Furthermore, because the dataset is limited to aging and age-related illnesses, it may be biased towards older adults and fail to account for variations in disease appearance in younger or other life stages.

Interpolation, scaling, and geometric image revolution methods, which provide 95% accuracy, were used to resize images (Kamath *et al*., 2021). The cognitive assessment was used on 3309 different subjects from MRI Alzheimer's disease pictures, producing 89.99% accuracy for data cleaning purposes (Bhol, 2019).

An accuracy of 95% in image resizing was reported by (Kamath *et al*., 2021) using interpolation, scaling, and geometric image revolution techniques. The cognitive assessment was used to clean the data, producing 89.99% accuracy on, 3309 different participants from MRI Alzheimer's disease pictures. The input images are normalized using the scaling and N4 bias field correction techniques, (Habuza *et al*., 2022). To balance the dataset, techniques including horizontal flipping, width shift, height shift and scaling have also been used, (Tufail *et al*., 2020). For further processing, the input photos were reduced in size and made grayscale, (Haider *et al*., 2020). FSL-BET was used to remove the non-brain tissues (Balaji *et al*., 2023). The ROI was used to transform the input images to a grayscale with an accuracy of 91.67% (Sharma *et al*., 2021). Down sampling was used to extract sick areas from MRI scans and thus organized the RGB values into pixels (Arvesen, 2015). To enhance the quality of the images, preprocessing techniques such spatial filtering (Jain *et al*., 1995), image correction (Klein *et al*., 2010), Gaussian filter (Csernansky *et al*., 2002) and histogram equalization (Senthilkumaran and Thimmiaraja, 2014), have also been used lately, but these studies showed the computational results on a limited clinical trial.

**Table 3:** Online freely available datasets

| Ref# | Year of release | Datasets | Classes | No. of images |
|---|---|---|---|---|
| Marcus *et al.* (2007) | 2007 | OASIS | 3 | 2168 |
| Dayananda *et al*. (2022) | 2007 | IBSR | 1 | 18 |
| Keator *et al.* (2016) | 2015 | FBIRN | 2 | 310 |
| Jack *et al.* (2008) | 2004 | ADNI | 2 | 1821 |
| Yiğit and Işik (2020) | 2012 | MIRIAD | 3 | 416 |
| Iglesias and Sabuncu (2015) | 2014 | MICCAI | 1 | 35 |

An overview of the accuracies in existing preprocessing techniques is presented in Table (4).

### *Segmentation Methods*

Segmentation segregates areas of interest which is carried out by utilizing a variety of computational techniques and distinct methods which is quite challenging, (Zhang *et al.*, 2019). On PET and MRI images, level-set methods and fuzzy-c-mean clustering were used to remove artifacts, (Mohanty *et al.*, 2020; Allada *et al.*, 2023; Lanjewar *et al.*, 2023). The background/foreground region of brain MRI images was extracted using U-Net, (Fan *et al.*, 2021; Ghosh *et al.*, 2019).

The ADNI dataset was used to detect AD and MCI images using graph cut and canny filters, (Wolz *et al.*, 2010). Despite being a useful tool for studying Alzheimer's disease, the ADNI dataset contains a number of possible biases. A noteworthy constraint is the prevalence of North American individuals, which may limit the applicability of results to other demographics with distinct genetic, environmental or lifestyle elements. Furthermore, because the ADNI program largely focuses on early diagnosis and monitoring, the dataset is frequently biased towards those who are reasonably healthy or in the early stages of the disease. This could cause biases in how the disease progresses in cases when it is more advanced or in those who have coexisting diseases.

Another technique known as region growth was used to segment the brain. Alzheimer's disease disease's morphological local characteristics were segmented using the SegNet DL method, (Buvaneswari and Gayathri, 2021). The hippocampus area can be divided to identify Alzheimer's disease using the k-mean clustering and water-shed approach, (Holilah *et al.*, 2021). These were a few additional techniques for AD segmentation. Hippocampal, VGG-Net and Hybrid CNN-S are additional methods. Other methods include multiscale convolutional neural network (MSCNet), dense encoder-decoder-based framework, (Stricker *et al.*, 2022; Qasim Abbas *et al.*, 2023 and Mahmud *et al.*, 2024) enhance fully convolutional network, (Monteiro *et al.*, 2022), attentive border aware network, hierarchical k-means algorithms with level set methods, such as k-means clustering optimized through Firefly Algorithm (FFA), (Liang *et al.*, 2023), FCM-based segmentation. An overview of the current segmentation methods, their accuracies on various datasets is shown in Table (5).

**Table 4:** Overview of the existing preprocessing methods' accuracies

| Refs. # | Years | Methods | Preprocessing methods | Datasets | Accuracy (%) |
|---|---|---|---|---|---|
| Kamath *et al.* (2021) | 2021 | Convolution neutral network | Image interpolation, scaling and geometric image revolution | OASIS | 95 |
| Bhol (2019) | 2019 | Recursive feature elimination, logistic regression, random forest, gradient boosting, light GBM, XG boost, neural network | Data cleaning, imputation, balancing, scaling | Tabular data | 88.3 |
| Duc *et al*. (2020) | 2020 | FSL toolbox, 3D CNN | FSL toolbox | National dementia research center chosun university dataset | 85.27 |
| Gupta *et al*. (2020) | 2020 | FMRIprep, FNN | FMRIprep | ADNI | 81 |
| Lee *et al*. (2021) | 2021 | DPARSF toolbox, CNN, GCN | DPARSF toolbox | ADNI | 74.42 |
| Balaji *et al*. (2023) | 2023 | ACO, MFCM, CNN, LSTM, DNN, IAO | ACO algorithm | Kaggle AD dataset | 98.5 |

**Table 5:** Overview of the current DL segmentation methods on various datasets

| Refs. # | Years | Methods | Segmentation methods | Datasets | Accuracy (%) |
|---|---|---|---|---|---|
| Stricker *et al*. (2022) | 2022 | CircNet, lncRNA | Dense encoder-decoder-based framework | Gencode | 98.28 |
| Liang *et al*. (2023) | 2022 | Multi-scale Fusion Module, DRMNet | Distilled multi-residual Network (DMR-Net) | ADNI | 83.90 |
| Liu *et al*. (2020) | 2020 | 3D DenseNet, CNN | Multi-model deep learning framework, Hippocampal | ADNI | 87 |
| Mehmood *et al.* (2021) | 2021 | VGG | VGG-Net | ADNI | 83.70 |
| Sethi *et al*. (2022) | 2022 | CNN, SVM | Hybrid CNN SVM | ADNI, OASIS | 88 |
| Buvaneswari and Gayathri (2021) | 2021 | SegNet, ResNet | SegNet | ADNI | 95.00 |
| Carmo *et al*. (2021) | 2021 | HarP, Hippocampus segmentation | 2D multi-orientation approach | Hippocampus | 89.00 |
| Helaly *et al*. (2022) | 2022 | DC-GAN, DL-AHS, SHPT-Net, RESU-Net | DL-AHS (deep Learning Alzheimer's disease's hippocampus segmentation) | ADNI, NITRIC | 97.00 |
| Katabathula *et al.* (2021) | 2021 | 3D deep convolutional network model | Dense CNN2 | ADNI | 97.80 |
| Basheer *et al*. (2021) | 2021 | EICA, Skull stripped algorithm | Enhanced Independent component analysis (EICA) | OASIS, ADNI | 98.00 |
| Balaji *et al*. (2023) | 2023 | MFCM, ACO, CNN, LSTM | MFCM (Modify Fuzzy C-Mean), ACO (Ant-Colony Optimization) | AD-related two datasets (MRI, PET) | 98.05 |

### *Feature Extraction Techniques*

One of the challenging tasks is to extract and collect the best features, which begins by producing values (features) from a set of valuable measured data. This process aids in generalization and learning and it may sporadically improve human analyses. To extract useful information from vast amounts of data, dimension reduction techniques are used, also dimension reduction and feature extraction are integrated. Numerous feature extraction techniques were proposed by various researchers, including modified ABCD (Skouras *et al*., 2019), Local Binary Pattern (Sharif *et al*., 2020a; Amin *et al*., 2019), Local Vector Pattern, LTrP (Bharathi and Manimegalai, 2016), GLCM (Sivapriya *et al*., 2011), higher entropy value features with Principal Component Analysis (PCA) (Whitwell *et al*., 2017 and Amin *et al*., 2017) and Feature Fusion (Sharif *et al*., 2020b). Gabor filter (Aruna and Chitra, 2016; Amin *et al*., 2020), wavelet transforms, SIFT (Xiao *et al*., 2022), Partial Swarm Algorithm (PSO), (Sharif *et al*., 2020a), Long and Short Term Memory (LSTM) and HOG (Bansal *et al*., 2022), etc. A classification method is proposed by the authors in (Padmavathi *et al*., 2023) to predict Alzheimer's disease in four steps, i.e., feature-based techniques, dimensionality reductions and feature elicitation and selection. These techniques require considerable amount of time and multiple optimization stages. A generic feature extraction algorithm on raw data is shown in Algorithm 1.

**Algorithm 1:** Features extraction

```
Algorithm 1: Feature Extraction Algorithm
Input: Data Repositories
Output: Raw Data
procedure Feature extraction
   D_image ←——— (path)
     Initialize data = Count the number of images and initialize min/max values for the image width and height.
   While (!end of file)
       update min/max width and height.
       D1_image ←——— Print data()
   return D1
end procedure
```

The above pseudocode explicitly shows the traditional features' selection methods on a repository of raw data. Here, we have analyzed two multivariate techniques that employ alternative strategies to focus on the challenge of partial model size.

It is possible to identify ROIs that are sensitive to the progression of AD by looking through the features obtained from the first hidden layer. The input pattern, $x^*$ can be derived by Eq. 6:

$$x^*_{ij} = \frac{W^1{}_{ij}}{||W_2||} \tag{6}$$

The variance $D^m$ of all the same $x^m$ ROIS, can by computed by splitting the pattern $x$ into $m$ features. The extracted features are considered to be more stable for AD diagnosis when $D_j^m$ is low. Hence, an overall feature stability $S_j$ of $j^{th}$ ROI can be computed as Eq. 7:

$$S_j = \sum_m \frac{\sum_j D_j^m}{D_j^m} \tag{7}$$

$S$ can be convolved with a Gaussian filter for exaggerating the differences between each ROI. To exaggerate the differences between each, ROI'S can be convolved using a Gaussian filter. Following the estimation of the Gaussian mixture model using the expectation maximizing approach, the feature directions for each image are then determined based on the positions of the created Gaussian. The alternate approach under consideration creates the score vectors via the Partial Least Squares (PLS) method, which are subsequently used as features.

Where:

$$D = (X, Y) \tag{8}$$

A Bayesian probabilistic method for simulating the relationship between inputs x and outputs y is called GP-LR. Authors in (Challis *et al*., 2015) proposed a Gaussian-based classification of Alzheimer's disease before these automated classification algorithms can be used in the clinic, it is still necessary to distinguish between different illnesses, which was outside the focus of the study.

The feature extraction and classification methods are shown in Fig. (2).

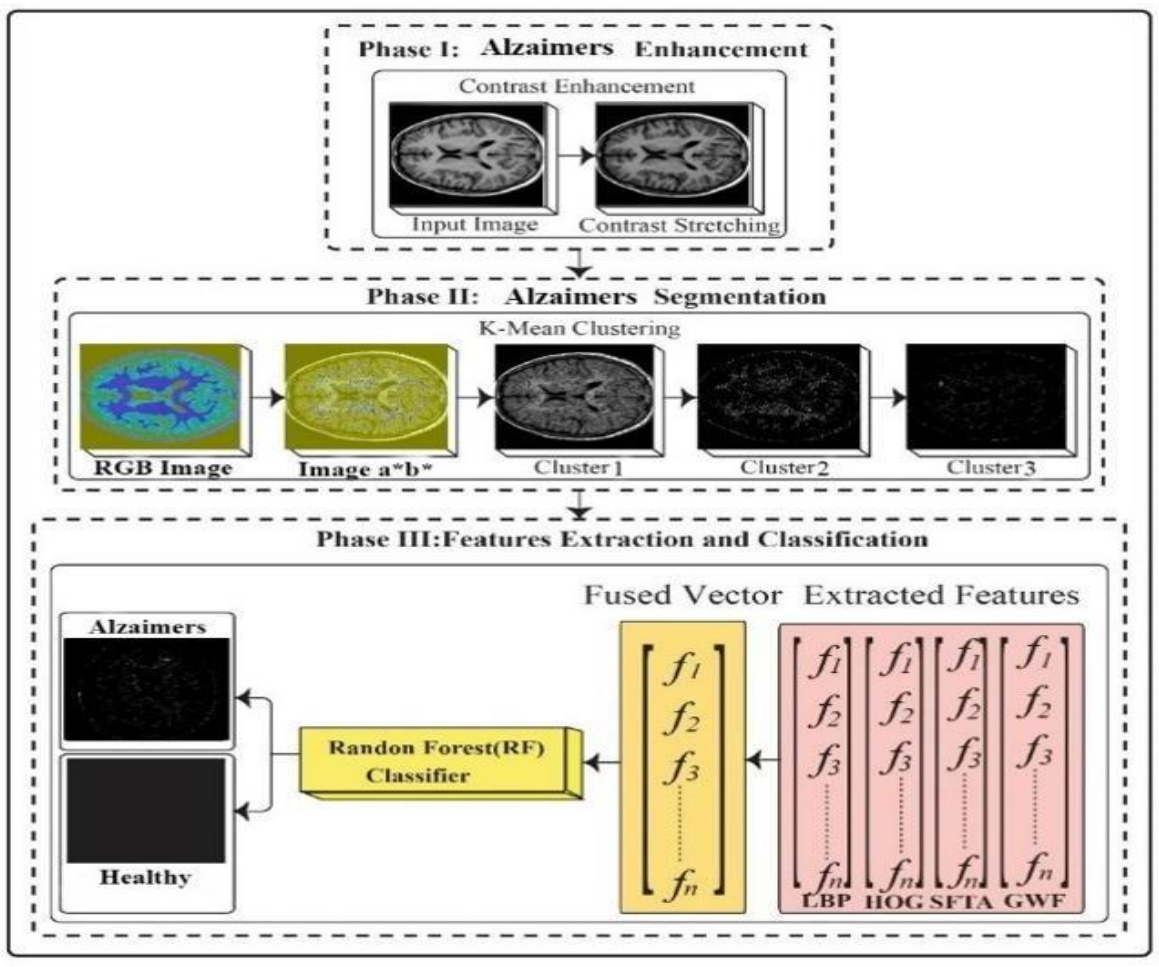

**Fig. 2:** Feature extraction and classification

The efficacy of these techniques is validated using the ADNI database by establishing multiple CAD approaches utilizing linear and nonlinear classifiers and comparing them to earlier tactics like Visual Attention Feature (VAF), (Segovia *et al*., 2012). This study focuses on Grey Matter (GM) pictures from all regions of the brain were split into 3D chunks using Automated Anatomical Labelling (AAL) mapping sections, resulting in multiple fully integrated systems. A large picture dataset from the ADNI was utilized to assess the methods. For Alzheimer's disease categorization, classification yields accuracy between 0.90 and 0.95. (Ortiz *et al*., 2016).

A multi-model deep Learning approach created on convolutional neural network for integrated automated hippocampal segmentation and Alzheimer's disease diagnosis is presented by the researchers using systemic MRI images as shown in Fig. (3).

Alzheimer's diseasestageT1-weighted benchmark structural MRI data from the ADNI database, which includes 119 Normal Control, 97 Alzheimer's disease and 233 MCI participants, is used to evaluate their method. To multitask, Convolutional Neural Network and DenseNet algorithms' modeled characteristics were combined to diagnose Alzheimer's disease condition. The recommended technique achieved an accuracy of 88S.9% and an AUC of 92.5% for diagnosing AD vs. NC subjects, (Liu *et al*., 2020). This approach employed MR images to identify MCI-to-AD change using a DL technique derived from CNN.

A convolutional neural network was trained using the regions from AD and NC to identify DL features in MCI patients. FreeSurfer was used to extract structural brain image features to assist CNN, (Amin *et al*., 2019). Finally, to predict the AD conversion, both kinds of characteristics were given to an ML classifier. The suggested method is proven using the ADNI MRI datasets. This method has a 79.9% accuracy, (Lin *et al*., 2018). To predict an Alzheimer's disease change, many different features are used in an EL or MC (Machine Classifier). Moreover, the integrated CNN and Extreme Learning technique can provide a distinct understanding of the complex variation of the entire MR image changes in Alzheimer's disease's, which will improve the ensemble model's classification capacity to identify the early-stage brain abnormalities connected with AD. Classification accuracy for MCI vs. Healthy Control is 0.79-0.04%, MCI vs. MCInc is 0.62, 0.06% and AD vs. HC is 0.84-0.05%, (Pan *et al*., 2020).

Figure (4) graphically represents the existing feature extraction techniques.

## *Classification*

The classification technique divides the input data into two classes (Alzheimer's disease and non- Alzheimer's disease's) as presented in the dataset. The classification algorithm is trained using labeled data and then later unlabeled data is fed into the learned classifier, which categorizes the data and returns results. For Alzheimer's disease classification, various classifiers such as KNN (Chudhey *et al*., 2022), SVM (Bharati *et al*., 2022), ensemble, DT (Faouri *et al*., 2022), decision tree-based random forest (Wang *et al*., 2022) and Naive Bayes (Khanna *et al*., 2022) were used. Deep learning models such as VGG16, VGG19 (Antony *et al*., 2023), MobileNet (Kumar *et al*., 2025), GoogleNet (Khanna *et al*., 2022), Inception-ResNet-v2 (Bhardwaj *et al*., 2022), ResNet-50 (Sethi and Ahuja, 2023) and Inception v3 (Sakatani and Yener, 2022), were also used for categorization. The researcher proposed a CNN DL architecture in this study. The researcher employed the MMSE and APOE4 as biomarkers to improve the Alzheimer's disease analysis and the accuracy is 92.89% (Thibeau-Sutre *et al*., 2022). Convolution neural networks have lately attracted scientific attention due to their exceptional performance in the processing and categorization of MRI images. The classification of AD using CNN model is shows in Fig. (5).

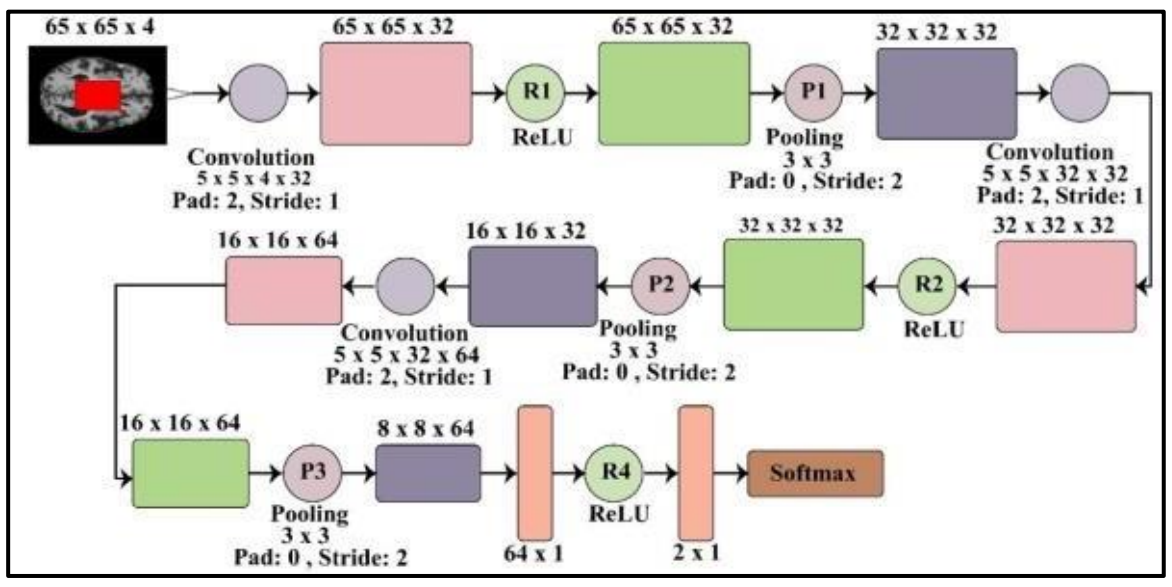

**Fig. 3:** Alzheimer's disease classification via MRI images (Amin *et al*., 2018)

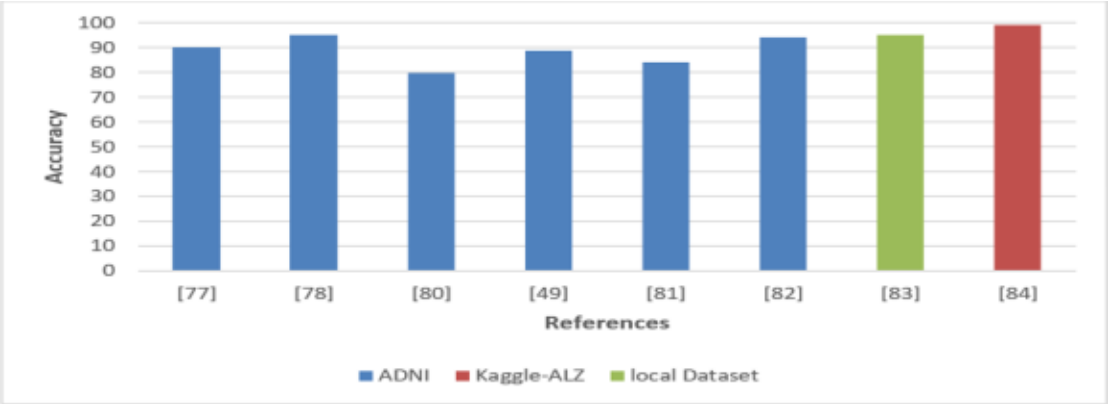

**Fig. 4:** Feature extraction accuracy using different techniques

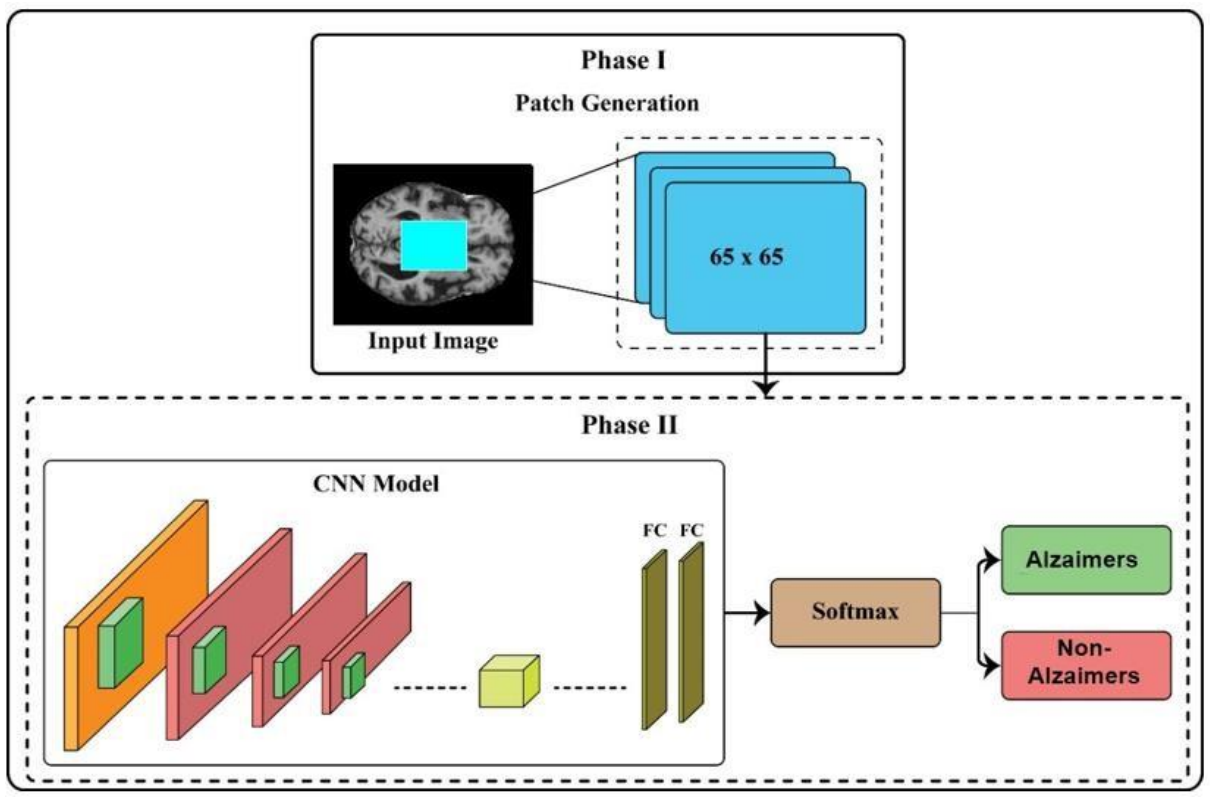

**Fig. 5:** Classification of AD using the CNN model (Amin *et al*., 2019)

However, there are significant demerits to CNN, such as the massive dataset required for training. Resulting, in Inception-ResNet V2 Convolutional Neural Network approach merged with the learning algorithm. The dataset was accessed via the ADNI database. The validation of a five- way classification yielded an accuracy of 97.3% (Bhardwaj *et al*., 2022). The super pixel-powered autoencoder technique, proposed in (Amin *et al*., 2019), recovers the key features using a histogram of targeted gradients. The proposed technique can identify and classify three forms of AD: Healthy, MCI and Alzheimer's disease patients. Its usefulness has been demonstrated by comparing the suggested strategy to current ad-vanced methods and well-trained algorithms such as VGG19, Resnet50, Inceptionv3, SqueezeNet, Resnet18, GoogleNet and Alexnet (Amin *et al*., 2019). (Kumar *et al*., 2025) used the ResNet50, Inception V3, Inception ResNet V2, DenseNet and MobileNet architectures for this classification. We assessed the performance of each design and selected the one that appeared to be the most promising. To achieve promising results, learning-based systems such as ResNet 50, which had previously been trained on ImageNet data, were modified to use ADNI data on a variety of input variables. The four optimizers used are SGD, Adagrad, Rmsprop and Adam, with two distinct set sizes. When tested with batches of 16 and 32 using the SGD and Adam optimizers, the results show that the optimizers' rmsProp and Adamax functioned satisfactorily.

For batch size 32, the classification accuracy for Alzheimer's disease vs. Normal Control using Rmsprop is 74.22%. Using batch size 16 resulted in a moderate improvement of 1.01% and classification accuracy of 75.6% (Sethi and Ahuja, 2023). DL algorithms such VGG-16 and VGG-19 DL(CNN) were used in this study to examine the precision of cognitively normal vs MCI, cognitively normal vs AD and MCI to Alzheimer's disease change utilizing MR imaging data. The suggested model examined and tested ADNI MR image data (Antony *et al*., 2023). An artificial neural classifier, also known as a boosting classifier, was deliberately coupled with all the previous classifiers to achieve the highest accuracy. The results showed a 20% improvement over cutting-edge machine learning algorithms, with a maximum accuracy of 93.33% (Chudhey *et al*., 2022). Table (6) presents the summary of existing classification methods.

A wide variety of datasets from many different fields, including medical imaging and diagnostics, are available on Kaggle. On the other hand, Kaggle datasets may display a number of biases.

Mild Cognitive Impairment (MCI) can be a precursor to dementia, especially in Alzheimer's disease (Gauthier *et al*., 2006; Davatzikos *et al*., 2008; Klöppel *et al*., 2009; Lao *et al*., 2004; Sarica *et al*., 2017; Jo *et al*., 2019).

### *Performance Measures*

The detection performance of Alzheimer's disease and non-Alzheimer's disease is mostly defined by performance metrics. To compare how Alzheimer's disease and non-Alzheimer's disease perform, we use the metrics of Specificity, Sensitivity, Accuracy, Precision, Positive and Negative Prediction Value and Area Under the Curve. Sensitivity and specificity improve and the ratio of dementia detection.

### *Accuracy*

Equation (8) is used to validate the accuracy:

$$ACC = \frac{TP+TN}{TP+FP+TN+FN} \quad (8)$$

### *Specificity*

Specificity is measured by calculating the ratio of True Negative to False Positive, as shown in Eq. (9):

$$FBR = \frac{TN}{TN+FP} \quad (9)$$

### *Sensitivity*

The next metric is sensitivity also known as Recall, which shows the classifier's ability to locate all positive samples as shown in Eq. (10):

$$TPR = \frac{TP}{TN+FP} \tag{10}$$

### *Area Under the Curve*

This metric is measured by computing sensitivity as shown in Eq. (11):

$$AUC = \int_{\infty}^{-\infty} TRP(T)FPR(T)dt \tag{11}$$

**Table 6:** Summary of the existing classification techniques

| Refs. # | Years | Techniques | Datasets | Results (%) |
|---|---|---|---|---|
| Bharathi and Manimegalai (2016) | 2022 | Random forest, XGBoost, voting-based, gradient boosting | OASIS | 92.00 |
| Yiğit *et al.* (2022) | 2022 | Ensemble deep neural networks | NIfTI | 85.00 |
| Basheer *et al.* (2021) | 2021 | CapsNet | OASIS | 90.00 |
| EL-Geneedy *et al.* (2023) | 2023 | MFCM, ACO, CNN, LSTM | Kaggle (MRI, PET) | 95.00 |
| Balaji *et al.* (2023) | 2023 | DenseNet121, ResNet50, VGG16, EfficientNetB7, InceptionV3 | ADNI | 96.00 |
| EL-Geneedy *et al.* (2023) | 2023 | CNN, CAD-ALZ | Kaggle-ALZ | 98.00 |

## Research Findings and Discussion

In Gray *et al.* (2011), the authors proposed an automated CAD system using several soft computing concepts. Each strategy has advantages and disadvantages of its own.

However, none of the strategies produce predictable results. In light of this, this research seeks to give a comprehensive assessment of the present current techniques suggested in diagnosing AD, (Lei *et al*., 2022). It is noticed that in the literature, the objective is to categorize brain pictures and only a few approaches employ the information from ROIs (Lei *et al*., 2022). However, these approaches require more computing time, (Lin *et al*., 2021).

Variations in brain pictures make detecting Alzheimer's disease more difficult. Alzheimer's disease is difficult to diagnose due to the different structures of the human brain caused by age, gender and sickness categories. There are still many challenges with the complex visual elements of brain imaging, indicating that there is still room for improvement. The conclusions of the extant literature are as follows:

- Morphology, sex, sickness and age all create significant changes in brain architecture. It is difficult to employ a single segmentation strategy across all phenotypic classes for consistent efficiency (Kumar *et al*., 2018)
- The anatomical structure's contrast enhancement in T1, T2 and FLAIR modalities causes poor segmentation results (Zollanvari *et al*., 2021)
- The noisy backdrop of a standard image makes segmentation challenging since it is difficult to precisely identify each pixel or cell with known properties
- The compact size and volume of the brain region, one of the important predictors of AD, as well as its structural variation, contrast enhancement variations, low resolution, low signal-to-noise ratio and ambiguous border make segmentation difficult (Yoon and Yoon, 2012)
- Deep neural network ensembles were unable to classify dementia symptoms in a binary manner. As a result, the unique feature selection technique must be modified (Yiğit and Işik, 2020)
- Because the anatomical structure in MRI has poor contrast, identifying Alzheimer's disease automatically is difficult. Due to varied scanning conditions, the classification accuracy may also be reduced if noisy or outlier pixels are present in MRI pictures
- The biggest challenge with Alzheimer's disease is its difficulty observing and studying the patient's patho-physiology over time. There are only 152 alterations in total among the 2731 MRIs in the Alzheimer's disease neuro-imaging project dataset (Amin *et al*., 2019)

Table (7) provides an overview of various gaps and limitations in the existing approaches.

**Table 7:** Summary of the existing techniques with limitations

| Refs. # | Years | Datasets | Techniques | Research/Limitations |
|---|---|---|---|---|
| Bharathi and Manimegalai (2016) | 2022 | OASIS | Group of various ML techniques (RF, XGB, voting-based, gradient boosting) | Optimum features extraction and selection method is required to increase the multi-dementia classification |
| Yiğit *et al.* (2022) | 2022 | NIfTI | Ensemble deep neural networks | Ensemble DNN failed for binary classification of dementia disease; therefore, needs improvement for the novel features selection approach |
| Basheer *et al.* (2021) | 2021 | OASIS | CapsNet | MRI images are noisy and still needs preprocessing technique to improve image quality |
| EL-Geneedy *et al.* (2023) | 2023 | Kaggle (MRI, PET) | MFCM, ACO, CNN, LSTM | Enhance more accuracy with deep CNN algorithms |
| Balaji *et al.* (2023) | 2023 | ADNI | DenseNet121, ResNet50, VGG16, EfficientNetB7, InceptionV3 | Using data mining to enhance performance And efficacy of Alzheimer's disease prediction in the initial stages |
| EL-Geneedy *et al.* (2023) | 2023 | Kaggle-ALZ | CNN, CAD-ALZ | Heavy software and hardware for implementation |

## Conclusion

In reference to brain segmentation and Alzheimer's disease categorization, the growth of artificial intelligence, deep learning techniques and machine recognition in MRI imaging have sparked a lot of interest. The state of deep learning-based AD detection is reviewed in this study. We outlined the most recent findings and trends by a thorough examination of the literature, encompassing over 100 papers. In particular, we discuss the various approaches to handle neuroimaging data from single-modality and multi-modality investigations, the necessary preprocessing processes and relevant biomarkers and features (personal information, genetic data and brain scans). A detailed description of the performance of deep models is given. Identifying a trustworthy, general method has always been challenging, even if deep learning techniques have had a considerable influence on the mathematical analysis of Alzheimer's disease MRI. The efficiency of deep learning approaches may be influenced by preprocessing, implementation and post-processing. Researchers have presented an overview of these techniques, but these studies showed the computational results of a limited clinical trial. Furthermore, we have investigated how segmenting brain structure improves Alzheimer's disease classification accuracy. Brain MRI data segmentation can be problematic be- cause of the images' poor contrast, fuzzy backgrounds and contrast enhancement impact. However, due to the dispersion of morphological markers in brain MRI, automated Alzheimer's disease categorization is problematic. The most current breakthroughs in machine learning have been reviewed, including the sorts of data used and the accuracy with which machine learning algorithms can identify early-stage Alzheimer's diseases. Machine learning, by definition, improves prediction accuracy. Accuracy ranged between 80-98% when using convolutional neural networks and 3D CNNs. Nevertheless, various techniques still require some improvement, but the results are encouraging and this solution is determined to be a significant tool to assist physicians and other personnel in healthcare.

## Acknowledgment

We would like to thank Dr. Javeria Amin, Associate Professor University of Wah, for her advice, motivation, input and support during the creation of the manuscript.

## Funding Information

The authors have not received any financial support or funding to report.

## Author's Contributions

**Rubab Hafeez:** Conceptualization, methodology, software design.
**Sadia Waheed:** Data curation, written-original draft preparation, interim review, and editing.
**Syeda Aleena Naqvi:** Review, and editing.
**Fahad Maqbool:** Review, and editing. Visualization, investigation.
**Amna Sarwar:** Visualization, investigation.
**Sajjad Saleem:** Written-reviewed.
**Muhammad Imran Sharif, Kamran Siddique and Zahid Akhtar:** Supervision and written-reviewed and finalization.

## Ethics

This research does not involve any human participants, personal data, or ethical concerns related to privacy, bias, or harm. This work is not submitted or published anywhere else.

## References

Agarap, A. F. (2018). Deep Learning Using Rectified Linear Units (ReLu). *ArXiv:1803.08375, 1*.

Ahirwar, A. (2013). Study of Techniques Used for Medical Image Segmentation and Computation of Statistical Test for Region Classification of Brain MRI. *International Journal of Information Technology and Computer Science*, *5*(5), 44–53. https://doi.org/10.5815/ijitcs.2013.05.06

Allada, A., Bhavani, R., Chaduvula, K., & Priya, R. (2023). Early Diagnosis of Alzheimer's Disease From MRI Using Deep Learning Models. *Journal of Information Technology Management*, *15*(Special Issue), 52–71. https://doi.org/10.22059/jitm.2022.89411

Amin, J., Sharif, A., Gul, N., Anjum, M. A., Nisar, M. W., Azam, F., & Bukhari, S. A. C. (2020). Integrated Design of Deep Features Fusion for Localization and Classification of Skin Cancer. *Pattern Recognition Letters*, *131*, 63–70. https://doi.org/10.1016/j.patrec.2019.11.042

Amin, J., Sharif, M., Raza, M., Saba, T., & Anjum, M. A. (2019). Brain Tumor Detection Using Statistical and Machine Learning Method. *Computer Methods and Programs in Biomedicine*, *177*, 69–79. https://doi.org/10.1016/j.cmpb.2019.05.015

Amin, J., Sharif, M., Yasmin, M., & Fernandes, S. L. (2018). Big Data Analysis for Brain Tumor Detection: Deep Convolutional Neural Networks. *Future Generation Computer Systems*, *87*, 290–297. https://doi.org/10.1016/j.future.2018.04.065

Amin, J., Sharif, M., Yasmin, M., Ali, H., & Fernandes, S. L. (2017). A Method for the Detection and Classification of Diabetic Retinopathy Using Structural Predictors of Bright Lesions. *Journal of Computational Science*, *19*, 153–164. https://doi.org/10.1016/j.jocs.2017.01.002

Antony, F., Anita, H. B., & George, J. A. (2023). Classification on Alzheimer's Disease MRI Images with VGG-16 and VGG-19. *IOT with Smart Systems*, *312*, 199–207. https://doi.org/10.1007/978-981-19-3575-6_22

Arafa, D. A., Moustafa, H. E.-D., Ali, H. A., Ali-Eldin, A. M. T., & Saraya, S. F. (2024). A Deep Learning Framework for Early Diagnosis of Alzheimer's Disease on MRI Images. *Multimedia Tools and Applications*, *83*(2), 3767–3799. https://doi.org/10.1007/s11042-023-15738-7

Aruna, S. K., & Chitra, S. (2016). Machine Learning Approach for Identifying Dementia from MRI Images. *International Journal of Computer and Information Engineering*, *9*(3), 881–888.

Arvesen, E. (2015). *Automatic Classification of Alzheimer's diseasefrom Structural MRI*.

Ávila-Jiménez, J. L., Cantón-Habas, V., Carrera-González, M. del P., Rich-Ruiz, M., & Ventura, S. (2024). A Deep Learning Model for Alzheimer's Disease Diagnosis Based on Patient Clinical Records. *Computers in Biology and Medicine*, *169*, 107814. https://doi.org/10.1016/j.compbiomed.2023.107814

Balaji, P., Chaurasia, M. A., Bilfaqih, S. M., Muniasamy, A., & Alsid, L. E. G. (2023). Hybridized Deep Learning Approach for Detecting Alzheimer's Disease. *Biomedicines*, *11*(1), 149. https://doi.org/10.3390/biomedicines11010149

Bansal, D., Khanna, K., Chhikara, R., Dua, R. K., & Malhotra, R. (2022). A Superpixel Powered Autoencoder Technique for Detecting Dementia. *Expert Systems*, *39*(5), e12926. https://doi.org/10.1111/exsy.12926

Basheer, S., Bhatia, S., & Sakri, S. B. (2021). Computational Modeling of Dementia Prediction Using Deep Neural Network: Analysis on OASIS Dataset. *IEEE Access*, *9*, 42449–42462. https://doi.org/10.1109/access.2021.3066213

Belam, G., & Nilforooshan, R. (2021). The Use of Artificial Intelligence and Machine Learning in the Care of People with Dementia: A Literature Review. *European Psychiatry*, *64*(S1), S429–S429. https://doi.org/10.1192/j.eurpsy.2021.1144

Bharathi, A. S., & Manimegalai, D. (2016). Regional Atrophy Analysis of Alzheimer Brain Magnetic Resonance Images Using Local Texture Patterns. *ARPN Journal of Engineering and Applied Sciences*, *11*(1), 474–489.

Bharati, S., Podder, P., Thanh, D. N. H., & Prasath, V. B. S. (2022). Dementia Classification Using MR Imaging and Clinical Data with Voting Based Machine Learning Models. *Multimedia Tools and Applications*, *81*(18), 25971–25992. https://doi.org/10.1007/s11042-022-12754-x

Bhardwaj, S., Kaushik, T., Bisht, M., Gupta, P., & Mundra, S. (2022). Detection of Alzheimer Disease Using Machine Learning. *Smart Systems: Innovations in Computing*, *235*, 443–450. https://doi.org/10.1007/978-981-16-2877-1_40

Bhol, S. (2019). *Comparative analysis for the detection of Alzheimer's diseaseusing multiple machine learning models*.

Blom, E. S., Giedraitis, V., Zetterberg, H., Fukumoto, H., Blennow, K., Hyman, B. T., Irizarry, M. C., Wahlund, L.-O., Lannfelt, L., & Ingelsson, M. (2009). Rapid Progression from Mild Cognitive

Impairment to Alzheimer's Disease in Subjects with Elevated Levels of Tau in Cerebrospinal Fluid and the APOE ε4/ε4 Genotype. *Dementia and Geriatric Cognitive Disorders*, *27*(5), 458–464. https://doi.org/10.1159/000216841

Buvaneswari, P. R., & Gayathri, R. (2021). Deep Learning-Based Segmentation in Classification of Alzheimer's Disease. *Arabian Journal for Science and Engineering*, *46*(6), 5373–5383. https://doi.org/10.1007/s13369-020-05193-z

Carmo, D., Silva, B., Yasuda, C., Rittner, L., & Lotufo, R. (2021). Hippocampus Segmentation on Epilepsy and Alzheimer's Disease Studies with Multiple Convolutional Neural Networks. *Heliyon*, *7*(2), e06226. https://doi.org/10.1016/j.heliyon.2021.e06226

Challis, E., Hurley, P., Serra, L., Bozzali, M., Oliver, S., & Cercignani, M. (2015). Gaussian Process Classification of Alzheimer's Disease and Mild Cognitive Impairment from Resting-State fMRI. *NeuroImage*, *112*, 232–243. https://doi.org/10.1016/j.neuroimage.2015.02.037

Chudhey, A. S., Jindal, H., Vats, A., & Varma, S. (2022). An Autonomous Dementia Prediction Method Using Various Machine Learning Models. *Advances in Data and Information Sciences*, *318*, 283–296. https://doi.org/10.1007/978-981-16-5689-7_25

Csernansky, J. G., Wang, L., Jones, D., Rastogi-Cruz, D., Posener, J. A., Heydebrand, G., Miller, J. P., & Miller, M. I. (2002). Hippocampal Deformities in Schizophrenia Characterized by High Dimensional Brain Mapping. *American Journal of Psychiatry*, *159*(12), 2000–2006. https://doi.org/10.1176/appi.ajp.159.12.2000

Davatzikos, C., Fan, Y., Wu, X., Shen, D., & Resnick, S. M. (2008). Detection of Prodromal Alzheimer's Disease Via Pattern Classification of Magnetic Resonance Imaging. *Neurobiology of Aging*, *29*(4), 514–523. https://doi.org/10.1016/j.neurobiolaging.2006.11.010

Dayananda, C., Choi, J. Y., & Lee, B. (2022). A Squeeze U-SegNet Architecture Based on Residual Convolution for Brain MRI Segmentation. *IEEE Access*, *10*, 52804–52817. https://doi.org/10.1109/access.2022.3175188

Duc, N. T., Ryu, S., Qureshi, M. N. I., Choi, M., Lee, K. H., & Lee, B. (2020). 3D-Deep Learning Based Automatic Diagnosis of Alzheimer's Disease with Joint MMSE Prediction Using Resting-State fMRI. *Neuroinformatics*, *18*(1), 71–86. https://doi.org/10.1007/s12021-019-09419-w

Ebrahimighahnavieh, A., Luo, S., & Chiong, R. (2020). Deep Learning to Detect Alzheimer's Disease From Neuroimaging: A Systematic Literature Review. *Computer Methods and Programs in Biomedicine*, *187*, 105242. https://doi.org/10.1016/j.cmpb.2019.105242

EL-Geneedy, M., Moustafa, H. E.-D., Khalifa, F., Khater, H., & AbdElhalim, E. (2023). An MRI-Based Deep Learning Approach for Accurate Detection of Alzheimer's Disease. *Alexandria Engineering Journal*, *63*, 211–221. https://doi.org/10.1016/j.aej.2022.07.062

Fan, Z., Li, J., Zhang, L., Zhu, G., Li, P., Lu, X., Shen, P., Shah, S. A. A., Bennamoun, M., Hua, T., & Wei, W. (2021). U-Net Based Analysis of MRI for Alzheimer's Disease Diagnosis. *Neural Computing and Applications*, *33*(20), 13587–13599. https://doi.org/10.1007/s00521-021-05983-y

Faouri, S., AlBashayreh, M., & Azzeh, M. (2022). Examining Stability of Machine Learning Methods for Predicting Dementia at Early Phases of the Disease. *Decision Science Letters*, *11*, 333–346. https://doi.org/10.5267/j.dsl.2022.1.005

Ganesh, D., Aparna, C., Royal, C. J., Vinay, D., Hima Sri, D. S., & Kumar, M. S. (2023). Implementation of Convolutional Neural Networks for Detection of Alzheimer's Disease. *BioGecko, A Journal for New Zealand Herpetology*, *12*(01).

Gauthier, S., Reisberg, B., Zaudig, M., Petersen, R. C., Ritchie, K., Broich, K., Belleville, S., Brodaty, H., Bennett, D., Chertkow, H., Cummings, J. L., de Leon, M., Feldman, H., Ganguli, M., Hampel, H., Scheltens, P., Tierney, M. C., Whitehouse, P., & Winblad, B. (2006). Mild Cognitive Impairment. *The Lancet*, *367*(9518), 1262–1270. https://doi.org/10.1016/s0140-6736(06)68542-5

Ghosh, S., Chandra, A., & Mudi, R. K. (2019). A Novel Fuzzy Pixel Intensity Correlation Based Segmentation Algorithm for Early Detection of Alzheimer's Disease. *Multimedia Tools and Applications*, *78*(9), 12465–12489. https://doi.org/10.1007/s11042-018-6773-z

Gray, K. R., Wolz, R., Keihaninejad, S., Heckemann, R. A., Aljabar, P., Hammers, A., & Rueckert, D. (2011). Regional Analysis of FDG-PET for Use in the Classification of Alzheimer'S Disease. *2011 IEEE International Symposium on Biomedical Imaging: From Nano to Macro*, 1082–1085. https://doi.org/10.1109/isbi.2011.5872589

Gupta, S., Rajapakse, J. C., & Welsch, R. E. (2020). Ambivert Degree Identifies Crucial Brain Functional Hubs and Improves Detection of Alzheimer's Disease and Autism Spectrum Disorder. *NeuroImage: Clinical*, *25*, 102186. https://doi.org/10.1016/j.nicl.2020.102186

Habuza, T., Zaki, N., Mohamed, E. A., & Statsenko, Y. (2022). Deviation From Model of Normal Aging in Alzheimer's Disease: Application of Deep Learning to Structural MRI Data and Cognitive Tests. *IEEE Access*, *10*, 53234–53249. https://doi.org/10.1109/access.2022.3174601

Hahnloser, R. H., Sarpeshkar, R., Mahowald, M. A., Douglas, R. J., & Seung, H. S. (2000). Digital selection and analogue amplification coexist in a cortex-inspired silicon circuit. *nature*, *405*(6789), 947-951.

Haider, F., de la Fuente, S., & Luz, S. (2020). An Assessment of Paralinguistic Acoustic Features for Detection of Alzheimer's Dementia in Spontaneous Speech. *IEEE Journal of Selected Topics in Signal Processing*, *14*(2), 272–281. https://doi.org/10.1109/jstsp.2019.2955022

Helaly, H. A., Badawy, M., & Haikal, A. Y. (2022). Deep Learning Approach for Early Detection of Alzheimer's Disease. *Cognitive Computation*, *14*(5), 1711–1727. https://doi.org/10.1007/s12559-021-09946-2

Holilah, D., Bustamam, A., & Sarwinda, D. (2021). Detection of Alzheimer's Disease with Segmentation Approach Using K-Means Clustering and Watershed Method of MRI Image. *Journal of Physics: Conference Series*, *1725*(1), 012009. https://doi.org/10.1088/1742-6596/1725/1/012009

Ibrahim, R., Ghnemat, R., & Abu Al-Haija, Q. (2023). Improving Alzheimer's Disease and Brain Tumor Detection Using Deep Learning with Particle Swarm Optimization. *AI*, *4*(3), 551–573. https://doi.org/10.3390/ai4030030

Iglesias, J. E., & Sabuncu, M. R. (2015). Multi-Atlas Segmentation of Biomedical Images: A Survey. *Medical Image Analysis*, *24*(1), 205–219. https://doi.org/10.1016/j.media.2015.06.012

Jo, T., Nho, K., & Saykin, A. J. (2019). Deep Learning in Alzheimer's Disease: Diagnostic Classification and Prognostic Prediction Using Neuroimaging Data. *Frontiers in Aging Neuroscience*, *11*, 220. https://doi.org/10.3389/fnagi.2019.00220

Jack, C. R., Bernstein, M. A., Fox, N. C., Thompson, P., Alexander, G., Harvey, D., Borowski, B., Britson, P. J., L. Whitwell, J., Ward, C., Dale, A. M., Felmlee, J. P., Gunter, J. L., Hill, D. L. G., Killiany, R., Schuff, N., Fox-Bosetti, S., Lin, C., Studholme, C., … Weiner, M. W. (2008). The Alzheimer's Disease Neuroimaging Initiative (ADNI): MRI Methods. *Journal of Magnetic Resonance Imaging*, *27*(4), 685–691. https://doi.org/10.1002/jmri.21049

Jain, R., Kasturi, R., & Schunck, B. G. (1995). *Machine Vision*. *5*.

Kalkan, H., Akkaya, U. M., Inal-Gültekin, G., & Sanchez-Perez, A. M. (2022). Prediction of Alzheimer's Disease by a Novel Image-Based Representation of Gene Expression. *Genes*, *13*(8), 1406. https://doi.org/10.3390/genes13081406

Kamath, D., Fathima, M. F., K. P, M., & Mohanchandra, K. (2021). Early Detection of Alzheimer's Disease Using Convolutional Neural Network Architecture. *International Journal of Artificial Intelligence*, *8*(2), 48–57. https://doi.org/10.36079/lamintang.ijai-0802.232

Katabathula, S., Wang, Q., & Xu, R. (2021). Predict Alzheimer's Disease Using Hippocampus MRI Data: A Lightweight 3D Deep Convolutional Network Model with Visual and Global Shape Representations. *Alzheimer's Research and Therapy*, *13*(1), 104. https://doi.org/10.1186/s13195-021-00837-0

Keator, D. B., van Erp, T. G. M., Turner, J. A., Glover, G. H., Mueller, B. A., Liu, T. T., Voyvodic, J. T., Rasmussen, J., Calhoun, V. D., Lee, H. J., Toga, A. W., McEwen, S., Ford, J. M., Mathalon, D. H., Diaz, M., O'Leary, D. S., Jeremy Bockholt, H., Gadde, S., Preda, A., … Potkin, S. G. (2016). The Function Biomedical Informatics Research Network Data Repository. *NeuroImage*, *124*, 1074–1079. https://doi.org/10.1016/j.neuroimage.2015.09.003

Keshri, S., Kumar, R., Kumar, D., Singhal, T., Giri, S., Sharma, I., & Vatsha, P. (2022). Insights of Artificial Intelligence in Brain Disorder with Evidence of Opportunity and Future Challenges. *Journal of Pharmaceutical Negative Results*, *13*(9), 10853–10867. https://doi.org/10.47750/pnr.2022.13.S09.1267

Khanna, K., Bansal, D., Chhikara, R., Dua, R. K., & Malhotra, R. (2022). Analysis of Univariate and Multivariate Filters Towards the Early Detection of Dementia. *Recent Advances in Computer Science and Communications*, *15*(4), 611–619. https://doi.org/10.2174/2666255813999200930163857

Klein, J. C., Eggers, C., Kalbe, E., Weisenbach, S., Hohmann, C., Vollmar, S., Baudrexel, S., Diederich, N. J., Heiss, W. D., & Hilker, R. (2010). Neurotransmitter Changes in Dementia with Lewy Bodies and Parkinson Disease Dementia in Vivo. *Neurology*, *74*(11), 885–892. https://doi.org/10.1212/wnl.0b013e3181d55f61

Klöppel, S., Chu, C., Tan, G. C., Draganski, B., Johnson, H., Paulsen, J. S., Kienzle, W., Tabrizi, S. J., Ashburner, J., Frackowiak, R. S. J., & PREDICT-HD Investigators of the Huntington Study Group. (2009). Automatic Detection of Preclinical Neurodegeneration. *Neurology*, *72*(5), 426–431. https://doi.org/10.1212/01.wnl.0000341768.28646.b6

Kumar, M. A. K., Sivasankar, A., Sabbineni, A., & Madhu, A. (2025). Prediction and classification of Alzheimer's disease disease severity using deep learning model. *ArXiv:2501.15293*.

Kumar, M. S., Azath, H., Velmurugan, A. K., Padmanaban, K., & Subbiah, M. (2023). Prediction of Alzheimer's Disease Using Hybrid Machine Learning Technique. *AIP Conference Proceedings*, *2523*(1), 020091. https://doi.org/10.1063/5.0110283

Kumar, P., Nagar, P., Arora, C., & Gupta, A. (2018). U-Segnet: Fully Convolutional Neural Network Based Automated Brain Tissue Segmentation Tool. *2018 25th IEEE International Conference on Image Processing (ICIP)*, 3503–3507. https://doi.org/10.1109/icip.2018.8451295

Lanjewar, M. G., Parab, J. S., & Shaikh, A. Y. (2023). Development of Framework by Combining CNN with KNN to Detect Alzheimer's Disease Using MRI Images. *Multimedia Tools and Applications*, *82*(8), 12699–12717. https://doi.org/10.1007/s11042-022-13935-4

Lao, Z., Shen, D., Xue, Z., Karacali, B., Resnick, S. M., & Davatzikos, C. (2004). Morphological Classification of Brains Via High-Dimensional Shape Transformations and Machine Learning Methods. *NeuroImage*, *21*(1), 46–57. https://doi.org/10.1016/j.neuroimage.2003.09.027

Lee, J., Ko, W., Kang, E., & Suk, H.-I. (2021). A Unified Framework for Personalized Regions Selection and Functional Relation Modeling for Early MCI Identification. *NeuroImage*, *236*, 118048. https://doi.org/10.1016/j.neuroimage.2021.118048

Lei, B., Liang, E., Yang, M., Yang, P., Zhou, F., Tan, E.-L., Lei, Y., Liu, C.-M., Wang, T., Xiao, X., & Wang, S. (2022). Predicting Clinical Scores for Alzheimer's Disease Based on Joint and Deep Learning. *Expert Systems with Applications*, *187*, 115966. https://doi.org/10.1016/j.eswa.2021.115966

Liang, X., Wang, Z., Chen, Z., & Song, X. (2023). Alzheimer's Disease Classification Using Distilled Multi-Residual Network. *Applied Intelligence*, *53*(10), 11934–11950. https://doi.org/10.1007/s10489-022-04084-0

Lin, E., Lin, C.-H., & Lane, H.-Y. (2021). Deep Learning with Neuroimaging and Genomics in Alzheimer's Disease. *International Journal of Molecular Sciences*, *22*(15), 7911. https://doi.org/10.3390/ijms22157911

Lin, W., Tong, T., Gao, Q., Guo, D., Du, X., Yang, Y., Guo, G., Xiao, M., Du, M., & Qu, X. (2018). Convolutional Neural Networks-Based MRI Image Analysis for the Alzheimer's Disease Prediction from Mild Cognitive Impairment. *Frontiers in Neuroscience*, *12*, 777. https://doi.org/10.3389/fnins.2018.00777

Littau, T. (2022). *Characterization of familial Alzheimer's diseaseusing unsupervised learning algorithms a data analysis of Alzheimer's diseasecases.*

Liu, M., Li, F., Yan, H., Wang, K., Ma, Y., Shen, L., & Xu, M. (2020). A Multi-Model Deep Convolutional Neural Network for Automatic Hippocampus Segmentation and Classification in Alzheimer's Disease. *NeuroImage*, *208*, 116459. https://doi.org/10.1016/j.neuroimage.2019.116459

Mahmud, T., Barua, K., Habiba, S. U., Sharmen, N., Hossain, M. S., & Andersson, K. (2024). An Explainable AI Paradigm for Alzheimer's Diagnosis Using Deep Transfer Learning. *Diagnostics*, *14*(3), 345. https://doi.org/10.3390/diagnostics14030345

Marcus, D. S., Wang, T. H., Parker, J., Csernansky, J. G., Morris, J. C., & Buckner, R. L. (2007). Open Access Series of Imaging Studies (OASIS): Cross-Sectional MRI Data in Young, Middle Aged, Nondemented and Demented Older Adults. *Journal of Cognitive Neuroscience*, *19*(9), 1498–1507. https://doi.org/10.1162/jocn.2007.19.9.1498

Mehmood, A., Yang, S., Feng, Z., Wang, M., Ahmad, A. S., Khan, R., Maqsood, M., & Yaqub, M. (2021). A Transfer Learning Approach for Early Diagnosis of Alzheimer's Disease on MRI Images. *Neuroscience*, *460*, 43–52. https://doi.org/10.1016/j.neuroscience.2021.01.002

Mohanty, S., Mohanty, S., & Pattnaik, P. K. (2020). Smart Healthcare Analytics: An Overview. *Smart Healthcare Analytics in IoT Enabled Environment*, *178*, 1–8. https://doi.org/10.1007/978-3-030-37551-5_1

Monteiro, J. C., Yokomichi, A. L. Y., de Carvalho Bovolato, A. L., Schelp, A. O., Ribeiro, S. J. L., Deffune, E., & Moraes, M. L. de. (2022). Alzheimer's Disease Diagnosis Based on Detection of Autoantibodies Against Aβ Using Aβ40 Peptide in Liposomes. *Clinica Chimica Acta*, *531*, 223–229. https://doi.org/10.1016/j.cca.2022.04.235

Muhammed Raees, P. C., & Thomas, V. (2021). Automated Detection of Alzheimer's Disease Using Deep Learning in MRI. *Journal of Physics: Conference Series*, *1921*(1), 012024. https://doi.org/10.1088/1742-6596/1921/1/012024

Ortiz, A., Munilla, J., Górriz, J. M., & Ramírez, J. (2016). Ensembles of Deep Learning Architectures for the Early Diagnosis of the Alzheimer's Disease. *International Journal of Neural Systems*, *26*(07), 1650025. https://doi.org/10.1142/s0129065716500258

Padmavathi, B., Deeksha, R., Darshitha, H., & Ashwath, B. (2023). Alzheimer's Disease Classification Using Deep Learning Technique. *Journal of Survey in Fisheries Sciences*, *10*(3S), 2854–2864.

Pallawi, S., & Singh, D. K. (2023). Study of Alzheimer's Disease Brain Impairment and Methods for Its Early Diagnosis: A Comprehensive Survey. *International Journal of Multimedia Information Retrieval*, *12*(1), 7. https://doi.org/10.1007/s13735-023-00271-y

Pan, D., Zeng, A., Jia, L., Huang, Y., Frizzell, T., & Song, X. (2020). Early Detection of Alzheimer's Disease Using Magnetic Resonance Imaging: A Novel Approach Combining Convolutional Neural Networks and Ensemble Learning. *Frontiers in Neuroscience*, *14*, 259. https://doi.org/10.3389/fnins.2020.00259

Pushpa, B. R., Amal, P. S., & Kamal, N. P. (2019). Detection and Stagewise Classification of Alzheimer's Disease Using Deep Learning Methods. *International Journal of Recent Technology and Engineering (IJRTE)*, *7*(5S3), 206–212.

Qasim Abbas, S., Chi, L., & Chen, Y.-P. P. (2023). Transformed Domain Convolutional Neural Network for Alzheimer's Disease Diagnosis Using Structural MRI. *Pattern Recognition*, *133*, 109031. https://doi.org/10.1016/j.patcog.2022.109031

Sakatani, K., & Yener, G. (2022). Application of Machine Learning in the Diagnosis of Dementia. *Frontiers in Neurology*, *13*, 860607. https://doi.org/10.3389/fneur.2022.860607

Sarica, A., Cerasa, A., & Quattrone, A. (2017). Random Forest Algorithm for the Classification of Neuroimaging Data in Alzheimer's Disease: A Systematic Review. *Frontiers in Aging Neuroscience*, *9*, 329. https://doi.org/10.3389/fnagi.2017.00329

Segovia, F., Górriz, J. M., Ramírez, J., Salas-Gonzalez, D., Álvarez, I. I., López, M. M., & Chaves, R. (2012). A Comparative Study of Feature Extraction Methods for the Diagnosis of Alzheimer's Disease Using the ADNI Database. *Neurocomputing*, *75*(1), 64–71. https://doi.org/10.1016/j.neucom.2011.03.050

Senthilkumaran, N., & Thimmiaraja, J. (2014). Histogram Equalization for Image Enhancement Using MRI Brain Images. *2014 World Congress on Computing and Communication Technologies*, 80–83. https://doi.org/10.1109/wccct.2014.45

Sethi, M., & Ahuja, S. (2023). Hyper Parameters Tuning ResNet-50 for Alzheimer's Disease Classification on Neuroimaging Data. *Artificial Intelligence on Medical Data*, *37*, 287–297. https://doi.org/10.1007/978-981-19-0151-5_25

Sethi, M., Rani, S., Singh, A., & Mazón, J. L. V. (2022). A CAD System for Alzheimer's Disease Classification Using Neuroimaging MRI 2D Slices. *Computational and Mathematical Methods in Medicine*, *2022*(1), 1–11. https://doi.org/10.1155/2022/8680737

Sharif, M., Amin, J., Raza, M., Yasmin, M., & Satapathy, S. C. (2020a). An Integrated Design of Particle Swarm Optimization (PSO) with Fusion of Features for Detection of Brain Tumor. *Pattern Recognition Letters*, *129*, 150–157. https://doi.org/10.1016/j.patrec.2019.11.017

Sharif, M., Amin, J., Nisar, M. W., Anjum, M. A., Muhammad, N., & Ali Shad, S. (2020b). A Unified Patch Based Method for Brain Tumor Detection Using Features Fusion. *Cognitive Systems Research*, *59*, 273–286. https://doi.org/10.1016/j.cogsys.2019.10.001

Sharma, A., Kaur, S., Memon, N., Jainul Fathima, A., Ray, S., & Bhatt, M. W. (2021). Alzheimer's Patients Detection Using Support Vector Machine (SVM) with Quantitative Analysis. *Neuroscience Informatics*, *1*(3), 100012. https://doi.org/10.1016/j.neuri.2021.100012

Sivapriya, T. R., Saravanan, V., & Ranjit Jeba Thangaiah, P. (2011). Texture Analysis of Brain MRI and Classification with BPN for the Diagnosis of Dementia. *Trends in Computer Science, Engineering and Information Technology*, *204*, 553–563. https://doi.org/10.1007/978-3-642-24043-0_56

Skouras, S., Falcon, C., Tucholka, A., Rami, L., Sanchez-Valle, R., Lladó, A., Gispert, J. D., & Molinuevo, J. L. (2019). Mechanisms of Functional Compensation, Delineated by Eigenvector Centrality Mapping, Across the Pathophysiological Continuum of Alzheimer's Disease. *NeuroImage: Clinical*, *22*, 101777. https://doi.org/10.1016/j.nicl.2019.101777

Stopschinski, B. E., Del Tredici, K., Estill-Terpack, S.-J., Ghebremedhin, E., Yu, F. F., Braak, H., & Diamond, M. I. (2021). Anatomic Survey of Seeding in Alzheimer's Disease Brains Reveals Unexpected Patterns. *Acta Neuropathologica Communications*, *9*(1), 164. https://doi.org/10.1186/s40478-021-01255-x

Stricker, M., Asim, M. N., Dengel, A., & Ahmed, S. (2022). CircNet: An Encoder–Decoder-Based Convolution Neural Network (CNN) for Circular RNA Identification. *Neural Computing and Applications*, *34*(14), 11441–11452. https://doi.org/10.1007/s00521-020-05673-1

Thibeau-Sutre, E., Couvy-Duchesne, B., Dormont, D., Colliot, O., & Burgos, N. (2022). MRI Field Strength Predicts Alzheimer's Disease: A Case Example of Bias in the ADNI Data Set. *2022 IEEE 19th International Symposium on Biomedical Imaging (ISBI)*, 1–4. https://doi.org/10.1109/isbi52829.2022.9761504

Tufail, A. B., Ma, Y.-K., & Zhang, Q.-N. (2020). Binary Classification of Alzheimer's Disease Using sMRI

Imaging Modality and Deep Learning. *Journal of Digital Imaging*, *33*(5), 1073–1090. https://doi.org/10.1007/s10278-019-00265-5

Wang, J., Wang, Z., Liu, N., Liu, C., Mao, C., Dong, L., Li, J., Huang, X., Lei, D., Chu, S., Wang, J., & Gao, J. (2022). Random Forest Model in the Diagnosis of Dementia Patients with Normal Mini-Mental State Examination Scores. *Journal of Personalized Medicine*, *12*(1), 37. https://doi.org/10.3390/jpm12010037

Whitwell, J. L., Graff-Radford, J., Singh, T. D., Drubach, D. A., Senjem, M. L., Spychalla, A. J., Tosakulwong, N., Lowe, V. J., & Josephs, K. A. (2017). 18F-FDG PET in Posterior Cortical Atrophy and Dementia with Lewy Bodies. *Journal of Nuclear Medicine*, *58*(4), 632–638. https://doi.org/10.2967/jnumed.116.179903

Wolz, R., Heckemann, R. A., Aljabar, P., Hajnal, J. V., Hammers, A., Lötjönen, J., & Rueckert, D. (2010). Measurement of Hippocampal Atrophy Using 4D Graph-Cut Segmentation: Application to ADNI. *NeuroImage*, *52*(1), 109–118. https://doi.org/10.1016/j.neuroimage.2010.04.006

Wulandari, P., Novitasari, D. C. R., & Asyhar, A. H. (2018). Identification of Alzheimer's Disease in MRI Data Using Discrete Wavelet Transform and Support Vector Machine. *Proceedings of the International Conference on Mathematics and Islam*, 198–204. https://doi.org/10.5220/0008519301980204

Xiao, X., Xu, T., Liu, H., Liu, X., Liao, X., Zhou, Y., Zhou, L., Wang, X., Zhu, Y., Yang, Q., Hao, X., Liu, Y., Jiang, H., Guo, J., Wang, J., Tang, B., Li, J., Shen, L., & Jiao, B. (2022). CYLD Variants Identified in Alzheimer's Disease and Frontotemporal Dementia Patients. *Annals of Clinical and Translational Neurology*, *9*(10), 1596–1601. https://doi.org/10.1002/acn3.51655

Yiğit, A., & Işik, Z. (2020). Applying Deep Learning Models to Structural MRI for Stage Prediction of Alzheimer's Disease. *Turkish Journal of Electrical Engineering and Computer Sciences*, *28*(1), 196–210. https://doi.org/10.3906/elk-1904-172

Yiğit, A., Baştanlar, Y., & Işık, Z. (2022). Dementia Diagnosis by Ensemble Deep Neural Networks Using FDG-PET Scans. *Signal, Image and Video Processing*, *16*(8), 2203–2210. https://doi.org/10.1007/s11760-022-02185-4

Yoon, S. M., & Yoon, G.-J. (2012). Compressive Matting. *Advances in Visual Computing*, *7432*, 145–154. https://doi.org/10.1007/978-3-642-33191-6_15

Zhang, S., Sun, F., Wang, N., Zhang, C., Yu, Q., Zhang, M., Babyn, P., & Zhong, H. (2019). Computer-Aided Diagnosis (CAD) of Pulmonary Nodule of Thoracic CT Image Using Transfer Learning. *Journal of Digital Imaging*, *32*(6), 995–1007. https://doi.org/10.1007/s10278-019-00204-4

Zollanvari, A., Kunanbayev, K., Akhavan Bitaghsir, S., & Bagheri, M. (2021). Transformer Fault Prognosis Using Deep Recurrent Neural Network Over Vibration Signals. *IEEE Transactions on Instrumentation and Measurement*, *70*, 1–11. https://doi.org/10.1109/tim.2020.3026497